\documentclass[manuscript, nonacm]{acmart}

\AtBeginDocument{%
  }

\setcopyright{acmlicensed}
\copyrightyear{2018}
\acmYear{2018}
\acmDOI{XXXXXXX.XXXXXXX}
\acmConference[Conference acronym 'XX]{Make sure to enter the correct
  conference title from your rights confirmation email}{June 03--05,
  2018}{Woodstock, NY}
\acmISBN{978-1-4503-XXXX-X/2018/06}

\renewcommand\footnotetextcopyrightpermission[1]{} 

\usepackage{tabularx}
\usepackage{multirow} 

\begin{document}

\title{Fairness-Driven LLM-based Causal Discovery with Active Learning and Dynamic Scoring}

\author{Khadija Zanna}
\email{khzanna@rice.edu}
\orcid{https://orcid.org/0000-0002-3553-303X}
\affiliation{%
  \institution{Rice University}
  \country{USA}
}

\author{Akane Sano}
\email{akane.sano@rice.edu}
\orcid{https://orcid.org/0000-0003-4484-8946}
\affiliation{%
  \institution{Rice University}
  \country{USA}}

\renewcommand{\shortauthors}{Khadija Zanna and Akane Sano}

\begin{abstract}
  Causal discovery (CD) plays a pivotal role in numerous scientific fields by clarifying the causal relationships that underlie phenomena observed in diverse disciplines. Despite significant advancements in CD algorithms that enhance bias and fairness analyses in machine learning, their application faces challenges due to the high computational demands and complexities of large-scale data. This paper introduces a framework that leverages Large Language Models (LLMs) for CD, utilizing a metadata-based approach akin to the reasoning processes of human experts. By shifting from pairwise queries to a more scalable breadth-first search (BFS) strategy, the number of required queries is reduced from quadratic to linear in terms of variable count, thereby addressing scalability concerns inherent in previous approaches. This method utilizes an Active Learning (AL) and a Dynamic Scoring Mechanism that prioritizes queries based on their potential information gain, combining mutual information, partial correlation, and LLM confidence scores to refine the causal graph more efficiently and accurately. This BFS query strategy reduces the required number of queries significantly, thereby addressing scalability concerns inherent in previous approaches. This study provides a more scalable and efficient solution for leveraging LLMs in fairness-driven CD, highlighting the effects of the different parameters on performance. We perform fairness analyses on the inferred causal graphs, identifying direct and indirect effects of sensitive attributes on outcomes. A comparison of these analyses against those from graphs produced by baseline methods highlights the importance of accurate causal graph construction in understanding bias and ensuring fairness in machine learning systems. 
  
\end{abstract}





\maketitle

\section{Introduction} \label{sec:introduction}

Causal discovery (CD) identifies causal relationships among system entities, with applications in various scientific domains like epidemiology, genetics, and economics \cite{kampani2024llm}. Recent advances in CD algorithms have opened new avenues for addressing bias and fairness in machine learning (ML). Bias in ML often arises from hidden confounders or unequal treatment of sensitive groups \cite{barocas2023fairness}. For example, a hiring algorithm might unfairly prioritize candidates based on biased historical data, leading to discrimination \cite{binns2018fairness}. CD helps identify such pathways, distinguishing genuine effects from biases introduced by confounders and ensuring decisions are based on causal factors rather than proxies. This approach provides a principled method to address bias by uncovering the underlying causal mechanisms \cite{makhlouf2020survey}. Unlike traditional methods relying on statistical correlations or predefined fairness metrics, CD disentangles direct effects, indirect effects, and spurious correlations \cite{miguel2016comparing}.

Due to this, the scientific community has shifted focus to studying causality in the context of fairness within AI and ML domains. Seminal work by Binkyte et al. \cite{binkyte2023causal} emphasizes CD's role in understanding fairness and analyzing causal structures to mitigate bias effectively. Research by Zuo et al. \cite{zuo2022counterfactual} and Wang et al. \cite{wang2023causal} further advances our understanding of counterfactual fairness and structural causal models for correcting biases without relying on predefined causal graphs.


Despite advancements, traditional CD algorithms face high computational and time complexities, especially with large datasets \cite{karpatne2022knowledge}. These algorithms struggle with managing the complexity of causal networks and incorporating domain expert knowledge systematically \cite{takayama2024integrating}. To overcome these challenges, many recent works have demonstrated the potential of large language models (LLM) in CD \cite{le2024multi, vashishtha2023causal, khatibi2024alcm, kampani2024llm}. LLMs access vast training data, allowing them to generate plausible explanations and reason on counterfactuals \cite{wang2022pinto, yao2024tree}. Inspired by these works, the proposed framework utilizes LLMs to respond to causal queries based on metadata without accessing numerical observations.

Prior research by Long et al. \cite{long2023can}, Kıcıman et al. \cite{kiciman2023causal}, and Choi et al. \cite{choi2022lmpriors} show that LLMs can determine causal relationships between pairs based solely on metadata. However, expanding this pairwise methodology to complete graph discovery introduces scalability challenges. Jiralerspong et al. \cite{jiralerspong2024efficient} proposed an LLM-based CD framework using a breadth-first search (BFS) approach to reduce query numbers. This method is more efficient than previous LLM-based methods but still requires improvements in computational efficiency.

Building upon this BFS method \cite{jiralerspong2024efficient}, our work proposes an enhanced approach incorporating Active Learning (AL) and a Dynamic Scoring Mechanism. The original BFS method treats all queries equally and relies on manual prompts. In contrast, the proposed method prioritizes the most informative queries by dynamically scoring variable pairs based on mutual information (MI), partial correlation (PC), and LLM confidence scores. A query history weighting mechanism prevents redundant exploration.

Furthermore, this work analyzes fairness from causal graphs generated by the proposed method from a real-world dataset. This includes analyzing direct and indirect effects of sensitive variables on outcomes, identifying mediators, and quantifying disproportionate contributions of causal pathways. By comparing the fairness insights derived from the proposed method with those from baseline approaches, we illustrate the impact of improved causal graph accuracy on fairness evaluations.


In summary, in this work, we make the following contributions to developing an efficiency and accuracy-enhanced method for CD-based fairness analyses.

\begin{itemize}
    \item Enhancement of the breadth-first search (BFS) CD method by integrating AL and a Dynamic Scoring Mechanism to improve the informativeness and efficiency of LLM queries.
    \item Analysis on the effects of different parameters, such as query scoring weights, thresholds, and LLM configurations. This provides valuable insights into optimizing LLMs for CD tasks.
    \item Fairness analysis through pathway analysis of various CD methods, including the proposed method to understand the impact of improved causal graph accuracy on fairness evaluations.
\end{itemize}



\vspace{-2mm}
\section{Related Work}

Traditional methods, optimization-based approaches, and recent advances using LLM have contributed significantly to the field of CD in ML and statistics \cite{long2025survey}.

Traditional methods, such as the Peter Clark (PC) algorithm \cite{spirtes1991algorithm} and the Greedy Equivalence Search (GES) \cite{meek1997graphical}, infer causal graphs using statistical independence tests or heuristic scoring. However, these methods often rely on strong assumptions like causal sufficiency and faithfulness, and struggle with scalability in large networks. 
Optimization-based approaches, such as NOTEARS \cite{zheng2018dags} and DAGMA \cite{bello2022dagma}, address some of these limitations by formulating CD as a continuous optimization problem. Despite their high accuracy, they are computationally expensive and depend on large sample sizes, making them less suitable for real-world scenarios.

LLMs have recently emerged as a promising alternative for CD by incorporating domain knowledge into the process. For example, Kiciman et al. \cite{kiciman2023causal} leverages LLMs to infer pairwise causal relationships using natural language descriptions of variables, constructing graphs without observational data. However, this approach faces scalability challenges due to its quadratic query complexity. Other works, such as \cite{vashishtha2023causal}, combine LLM-inferred causal orders with constraint-based algorithms like PC to improve edge orientation. Frameworks like ALCM \cite{khatibi2024alcm} and Multi-Agent CD \cite{le2024multi} extend these ideas by integrating LLMs into the entire discovery process, refining graphs through prompts and multi-agent reasoning. Kampani et al. \cite{kampani2024llm} take a hybrid approach, using LLMs to generate priors and refining them with methods like NOTEARS.

Takayama et al. \cite{takayama2024integrating} propose statistical causal prompting, where LLMs enhance statistical methods by providing priors or plausible causal orders. While robust, this one-time augmentation lacks iterative refinement. Similarly, Jiralerspong et al. \cite{jiralerspong2024efficient} introduce a BFS-based approach that efficiently constructs causal graphs by iteratively exploring relationships. Despite reducing the complexity of the query, the method relies on fixed heuristics and does not dynamically balance statistical evidence with LLM insights.

Our proposed method builds upon these foundations, particularly the BFS approach of Jiralerspong et al. \cite{jiralerspong2024efficient}, and introduces key innovations. Using an AL-based mechanism, it prioritizes variable pairs with the highest uncertainty or informativeness, significantly reducing query overhead. A weighted scoring mechanism dynamically balances statistical metrics (e.g., mutual information and correlation) with LLM judgments, enhancing robustness and reducing over-reliance on the model. These weights are optimized using Bayesian optimization, enabling adaptation to diverse datasets and network structures.

Furthermore, the proposed framework incorporates a query-efficient refinement loop, iteratively validating and adjusting the graph based on statistical evidence and LLM feedback. This dynamic feedback-driven approach improves performance and scalability, addressing limitations in existing methods while achieving superior performance across networks.

\vspace{-0.3cm}

\section{Methodology} \label{sec:Method}

\subsection{Breadth-First Search (BFS) for Causal Discovery}

The foundation of the proposed method is the BFS-based CD framework introduced by \cite{jiralerspong2024efficient}. This approach iteratively queries an LLM to infer causal relationships among variables in a data set and reduces the number of queries required since visiting each node once is sufficient to determine all edges originating from that node. Using this prompting approach, each query is approached as an opportunity for node expansion within a Breadth-First Search (BFS) algorithm, progressively building the causal graph as it is traversed using BFS. To effectively implement the BFS, it is essential to determine the sequence in which nodes are explored. For this purpose, the Directed Acyclic Graph (DAG) nature of causal graphs is leveraged, by using their topological ordering to guide the BFS traversal. The BFS method is structured as follows:

\begin{enumerate}
    \item Initialization stage: The algorithm begins by querying the LLM to identify independent variables that are not caused by any other variables. These variables form the starting points for graph exploration.
    \item Expansion stage: For each variable on the current frontier (independent variables or those already explored), the LLM is queried to determine which variables it causes within the dataset. A prompt describing the causal relationships inferred so far is constructed for each query.
    \item Graph Construction: As the algorithm progresses, causal relationships (edges) are added to a directed graph. Each new relationship is checked to ensure that the variables in the frontier are connected to previously unexplored nodes.
    \item Iterative expansion: The process iterates until all variables are explored, expanding the frontier with newly discovered causal relationships. The final output is a predicted causal graph in the form of an adjacency matrix.
\end{enumerate}

While effective, the BFS approach explores all variable relationships exhaustively, often querying uninformative or redundant pairs. This leads to inefficiencies, particularly in large or sparse graphs.

\subsection{Active Learning and Dynamic Scoring}

To address the inefficiencies of the BFS approach, an AL and a Dynamic Scoring Mechanism are introduced to transform the BFS into a more targeted and adaptive method.

\subsubsection{Dynamic Scoring Mechanism}

To prioritize which pairs of variables to query, the algorithm computes a dynamic score $S(x,y)$ for each unqueried pair of variables $(x,y)$:

\begin{equation}
\label{eq:dynamic_score}
    S(x, y) = w_{\text{stat}} \cdot \text{StatScore}(x, y) + w_{\text{conf}} \cdot \text{LLMConf}(x, y) + w_{\text{hist}} \cdot \text{HistScore}(x, y)
\end{equation}

Where:

\begin{itemize}
    \item $w_{\text{stat}},w_{\text{conf}},w_{\text{hist}}$: Weights assigned to the statistical, confidence, and query history components.
    
    \item \textbf{Statistical score:}
    \begin{equation}
    \label{eq:stat_score}
        \text{StatScore}(x, y) = \frac{\text{MI}(x, y) + \text{PC}(x, y)}{2}
    \end{equation}
    where $\text{MI}(x, y)$ is mutual information and $\text{PC}(x, y)$ is partial correlation between $x$ and $y$.

    \item \textbf{LLM Confidence:}
    \begin{equation}
    \label{eq:LLM_conf}
        \text{LLMConf}(x, y) = \frac{1}{1 + e^{-\text{confidence}}}
    \end{equation}

    \item \textbf{Query History Score:}
    \begin{equation}
     \label{eq:hist_score}   
        \text{HistScore}(x, y) = \frac{1.5}{1 + \text{query\_count}(x, y)}
    \end{equation}
\end{itemize}

Equation \ref{eq:dynamic_score} combines multiple sources of information to prioritize the most informative variable pairs for querying. Each component is weighted to reflect its importance, allowing the algorithm to adapt its exploration strategy dynamically.

The statistical score in Equation \ref{eq:stat_score} averages mutual information (MI) and partial correlation (PC). MI captures both linear and non-linear dependencies \cite{song2012comparison}, while PC measures linear relationships controlling for other variables \cite{waliczek1996primer}. Averaging these metrics leverages their strengths, ensuring sensitivity to various relationships while remaining robust against confounding.

The LLM confidence score in Equation \ref{eq:LLM_conf} reflects the model's certainty in its causal judgments, derived from the probabilities of response tokens \cite{gligoric2024can}. High-confidence responses are mapped to values closer to 1, while low-confidence responses are closer to 0.5. If confidence information is unavailable, the score defaults to 0.5, indicating neutral confidence.

The query history score in Equation \ref{eq:hist_score} prioritizes unqueried or rarely queried variable pairs, with a maximum score of 1.5 for unqueried pairs. As a pair is queried more frequently, the score decreases, discouraging redundant queries and promoting balanced exploration of the variable space.

\subsubsection{Active Learning for Query Optimization}

\begin{figure}[h]
  \centering
  \includegraphics[scale=0.3]{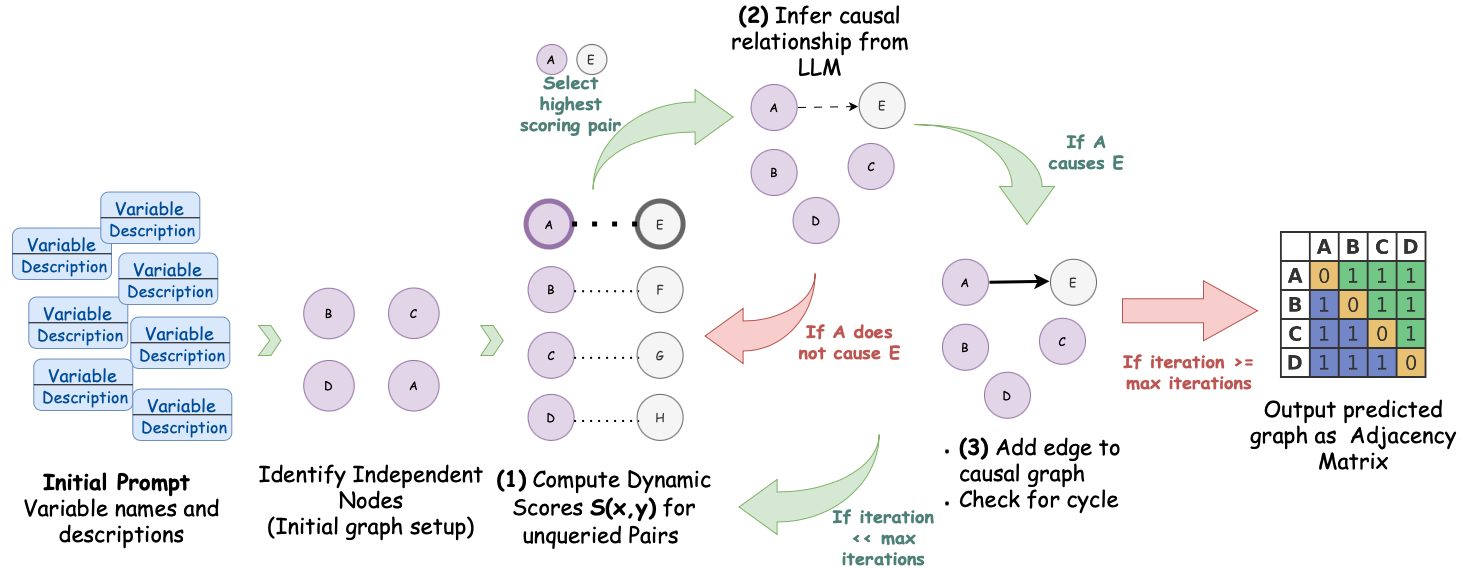}
  \caption{Proposed BFS and Active Learning Cycle Process}
  \label{fig:method_figure}
  \Description{}
\vskip -0.1in
\end{figure}

The BFS framework is augmented with AL to prioritize the most informative queries. Instead of querying all variable pairs, the informativeness of each pair is evaluated using the dynamic scoring mechanism and the pair with the highest score is queried. This targeted approach reduces redundant queries and focuses the LLM's attention on refining the causal graph. This process is visualized in Figure \ref{fig:method_figure}.

The AL framework iteratively selects the highest scoring variable pair $(x^*,y^*)$ and queries the LLM for a causal judgment. The loop proceeds as follows:

\begin{enumerate}
    \item Select Pair to maximize a dynamic score:
        \begin{equation}
            (x^*, y^*) = \arg\max_{(x, y) \in \text{UnqueriedPairs}} S(x, y)
        \end{equation}

    \item Query the LLM: Construct a natural language prompt:
    \begin{verbatim}
        Does X cause Y? Provide the answer in the format:
        <Answer>Yes</Answer> or <Answer>No</Answer>.
    \end{verbatim}

    \item Update graph: \begin{itemize}
        \item If the LLM response indicates causation and the addition does not create a cycle, add the edge $x^* \rightarrow y^*$.
        \item Update query history and confidence scores.
    \end{itemize}

    \item Termination Criteria: The loop stops if maximum iterations are reached or if top-scoring pairs fall below a threshold.

\end{enumerate}

To maintain the graph's DAG property, each edge addition is validated using cycle detection through the following steps:
\begin{itemize}
    \item Add the edge $x^* \rightarrow y^*$ to the graph.
    \item Check if the graph remains acyclic; if not, the edge is removed.
\end{itemize}

\begin{equation}
    A(i, j) = 
    \begin{cases} 
    1 & \text{if } X_i \rightarrow X_j \text{ is predicted}, \\
    0 & \text{otherwise}.
    \end{cases}
\end{equation}

The method outputs a causal graph as a dictionary and a predicted adjacency matrix A. The complete search process takes place within a single multi-turn chat environment. This enables the LLM to base each prompt on the preceding chat history.

\subsection{Complexity Analyses}

The original BFS method \cite{jiralerspong2024efficient} exhaustively explores variable relationships without prioritization, resulting in high computational complexity. It queries independent nodes with complexity $O(n)$, where $n$ is the number of variables. For each variable $x$, the LLM queries all potential children $y$, leading to a total query complexity of $O(n^2)$. Cycle prevention has complexity $O(n+e)$, where $e$ is the number of edges, making the worst-case overall complexity $O(n^3)$.

The proposed method introduces Active Learning (AL) and a Dynamic Scoring Mechanism to reduce the number of queries. Querying independent nodes remains $O(n)$. Pre-computing scores for all pairs $(x,y)$ has complexity $O(n^2)$. Dynamic scoring during each iteration involves updating scores, with complexity $O(k \cdot (n^2 - q))$, where $q$ is the number of queried pairs and $k$ is the maximum number of iterations. This approach prioritizes high-scoring pairs, reducing $q$ compared to the original BFS method. Cycle prevention complexity is $O(q' \cdot (n + e))$ for the proposed method, with $q' << n^2$. The overall complexity is $O(n^2 + k \cdot q' + q' \cdot (n + e))$, significantly improving efficiency.

Table \ref{tb:complexities} summarizes these computational complexity improvements.



\vskip -0.1in

\begin{table*}[h]
\caption{Computational Complexities of Proposed Method Vs. BFS Method}
\label{tb:complexities}
\begin{center}
\begin{scriptsize}
\begin{tabular}{|c|c|c|c|}
\hline
\textbf{Step} & \textbf{BFS method \cite{jiralerspong2024efficient}} & \textbf{Proposed Method} & \textbf{Improvement} \\
\hline
Independent Nodes & $O(n)$ & $O(n)$ & No improvement \\
\hline
Querying Relationships & $O(n^2)$ & $O(k \cdot q')$ & Reduces queries through prioritization. \\
\hline
Dynamic Scoring & Not applicable & $O(n^2 + k \cdot q')$ & Prioritizes high-impact queries. \\
\hline
Cycle Detection & $O(n^2 \cdot (n + e))$ & $O(q' \cdot (n + e))$ & Fewer edges evaluated due to early stopping. \\
\hline
Overall Complexity & $O(n^3)$ & $O(n^2 + k \cdot q')$ & Significant reduction in queries. \\

\hline
\end{tabular}
\end{scriptsize}
\end{center}
\vskip -0.1in
\end{table*}



In the worst-case scenario, where every variable pair must be queried, both methods approach $O(n^3)$. However, the proposed method is unlikely to reach the worst case due to low-priority pairs being avoided through the dynamic scoring mechanism, especially when they have negligible MI, PC, or confidence scores, and the threshold-based early stopping terminating uninformative queries. Through cycle prevention by rejecting edges that introduce cycles, the proposed method avoids redundant checks for dense graphs. Lastly, $q' << n^2$ in real-world data sets due to the sparsity in causal relationships and the efficiency of AL.

\subsection{Fairness Evaluation through Pathway Analyses}

To demonstrate the differences in fairness introduced by changes in the causal graph, pathway analyses are conducted on graphs produced by the proposed method and baseline methods from real-world datasets. Fairness in machine learning often requires understanding the direct and indirect effects of sensitive attributes (e.g., race, gender) on outcomes (e.g., income, hiring decisions) \cite{pearl2022direct}. Traditional fairness metrics, such as statistical parity or equalized odds, rely solely on observed associations, which may not account for confounding effects or hidden mediators. In contrast, causal pathway analyses leverage the discovered causal graph to provide a more nuanced assessment of fairness by disentangling direct effects from mediated or spurious pathways.

Using the causal graph discovered by the proposed LLM-based method, fairness evaluation focuses on identifying sensitive attributes and their causal relationships with key outcome variables. Specifically, the methodology examines: 


\begin{itemize}
    \item \textbf{Direct Paths} ($P_{\text{direct, fairness}}$): The number of direct causal edges ($S \rightarrow Y$) that connect the defined sensitive variables to the outcomes.
    \item \textbf{Indirect Paths} ($P_{\text{indirect, fairness}}$): The number of causal paths ($S \xrightarrow{} \text{intermediate nodes} \xrightarrow{} Y$) in which sensitive variables influence the outcomes indirectly via intermediate variables.
    \item \textbf{Spurious Paths} ($P_{\text{spurious, fairness}}$): The number of paths that involve sensitive variables but do not influence the outcomes ($S \xrightarrow{} \text{non-$Y$ nodes}$).
\end{itemize}




\section{Experiments}

To evaluate the proposed LLM-based CD framework with AL and dynamic scoring, a series of experiments are conducted on benchmark datasets. These experiments are designed to assess the effectiveness of the proposed method in improving the efficiency of causal graph discovery through iterative refinement and parameter optimization, and to understand the influence of the parameters. The details of the experimental setup, metrics, and results are outlined below.

\subsection{Datasets}

\subsubsection{Child Causal Network}

The "Child" causal graph is a medium-sized Bayesian network that simulates the causal relationships associated with congenital heart disease in newborns \cite{spiegelhalter1993bayesian}. The ground truth model of this graph comprises 20 nodes and 25 directed edges, representing a range of medical, parental, and environmental factors, as well as their interconnections that influence newborn health outcomes. The nodes in this graph represent clinical variables, including categorical and continuous types. Key variables in the graph include birth asphyxia (oxygen deprivation at birth), lung flow (airflow and lung capacity), chest X-ray results, cardiac mix (mixed blood flow levels), disease history, CVP (central vein pressure), and outcome measures such as mortality and length of stay in the hospital.

\subsubsection{Neuropathic Pain Causal Network}

The Neuropathic Pain Diagnosis Simulator, as detailed by Tu et al. \cite{tu2019Neuropathic}, uses a well-established causal graph that represents the pathophysiology of Neuropathic pain, with parameters estimated from actual patient data. This graph consists of 221 nodes and 770 edges, containing relationships between different nerves and the associated symptoms that patients express. The dataset includes symptoms diagnoses that describe patient discomfort, pattern diagnoses, and pathophysiological diagnoses. 

\subsubsection{UCI Adult}

The UCI Adult dataset \cite{kohavi1996scaling} is commonly used as a benchmark for studying fairness and bias in machine learning models. It consists of 48,842 data points, each described by 14 attributes and a binary target variable indicating whether an individual earns more than \$50,000 per year. The dataset includes information on employment, education, and demographics, with sensitive attributes being sex and race. 

\subsection{Experimental Design}

We validate the proposed method and study the influence of different Active Learning (AL) parameters on performance using Bayesian optimization with a Gaussian Process (GP) surrogate model. This approach balances exploration and exploitation, ensuring efficient optimization with minimal evaluations \cite{mockus1994application, snoek2012practical}. The scikit-optimize library's "gp\_minimize" function implements the GP-based Bayesian optimization.

The optimization space includes the following hyperparameters:

\begin{itemize}
    \item Weights: Proportions for MI and PC, correlation metrics, and query history in the scoring mechanism (defined in Equation \ref{eq:dynamic_score}).
    \item Score threshold: Minimum score to consider querying a variable pair.
    \item Temperature: LLM's temperature setting to control output variability.
    \item Maximum Iterations: Maximum number of AL algorithm iterations.
\end{itemize}

Each trial suggests a parameter configuration, and the CD framework executes these parameters. Implemented in Python, using the bnlearn library for Bayesian network handling and skopt for Bayesian optimization, with OpenAI GPT-4 for LLM queries.

To manage LLM API constraints and computational overhead, we use chunking to split the optimization process across multiple runs. Each chunk handles a fixed number of trials, with results saved for top-performing trials.

For the Child network, Bayesian optimization explores 1000 trials across all chunks, optimizing the graph F1 score compared to the ground truth. The F1 score balances precision and recall, providing a better indicator of edge prediction performance than accuracy. Parameter ranges include weights summing to 1, score thresholds from 0.01 to 0.2, temperature from 0.05 to 0.7, and maximum iterations from 10 to 50.

For the Neuropathic network, Bayesian optimization explores 200 trials across all chunks, optimizing the F1 score. Parameter ranges include weights summing to 1, score thresholds from 0.1 to 0.5, temperature from 0.1 to 0.3, to limit the randomness introduced into the model, and maximum iterations from 200 to 500, to account for the larger size of the network.

\subsection{Performance Evaluation Metrics}

To comprehensively evaluate the proposed method against the ground-truth graph, we use several metrics:

\paragraph{Precision}Proportion of correctly predicted causal edges among all edges predicted by the method. It is defined as:
$
\text{Precision} = \frac{\text{True Positives (TP)}}{\text{True Positives (TP)} + \text{False Positives (FP)}}
$

\paragraph{Recall} Proportion of true causal edges that were correctly identified by the method. 
$
\text{Recall} = \frac{\text{True Positives (TP)}}{\text{True Positives (TP)} + \text{False Negatives (FN)}}
$

\paragraph{Number of Predicted Edges vs. Number of True Edges}
Comparison between the total edge count in the predicted adjacency matrix and in the ground-truth adjacency matrix.

\paragraph{F1 Score}
Harmonic mean of precision and recall, offering a balanced metric for assessing the quality of edge predictions. 
{
$
\text{F1 Score} = 2 \cdot \frac{\text{Precision} \cdot \text{Recall}}{\text{Precision} + \text{Recall}}
$
}


\paragraph{DAG Validation}
Checks if both the ground-truth and predicted graphs are DAGs.

\paragraph{Normalized Hamming Distance (NHD)}
Quantifies the difference between the predicted adjacency matrix and the ground truth matrix.
$
\text{NHD} = \frac{\text{Number of mismatched entries}}{\text{Total entries in the adjacency matrix}}
$

\paragraph{Reference NHD}
Baseline metric calculated from a reference method to provide context to evaluate the performance of the proposed method.

\paragraph{Ratio}
 Compares NHD of the predicted graph to the reference NHD.
{
$
\text{Ratio} = \frac{\text{NHD (Predicted)}}{\text{NHD (Reference)}}
$
}

\paragraph{Accuracy}
Proportion of correctly predicted edges in the adjacency matrix.
{
$
\text{Accuracy} = \frac{\text{Correct Predictions}}{\text{Total Edges}}
$
}

\subsection{Baseline Methods}

The performance of the proposed method against the ground truth graph is compared with some baseline methods introduced in the following.

\begin{itemize}
    \item \textbf{Peter-Clark (PC) Algorithm}: This algorithm is a constraint-based method that uses conditional independence tests to iteratively prune edges from a fully connected graph, followed by orienting edges to form a DAG. It assumes causal sufficiency and the faithfulness assumption, relying heavily on accurate independence tests \cite{spirtes1991algorithm}.
    \item \textbf{Greedy Equivalence Search (GES) Algorithm}: GES is a score-based method that operates in two phases: a forward phase that incrementally adds edges to improve a score function (e.g., BIC), and a backward phase that removes edges to optimize the graph structure further. It assumes causal sufficiency and focuses on equivalence classes of DAGs to improve computational efficiency \cite{meek1997graphical}.
    \item \textbf{NOTEARS Algorithm}: NOTEARS formulates causal discovery as a continuous optimization problem by introducing a differentiable acyclicity constraint. This allows it to learn DAGs using gradient-based optimization, balancing data fit and sparsity \cite{zheng2018dags}.
    \item \textbf{DAGMA Algorithm}: DAGMA integrates neural networks and optimization techniques to model nonlinear causal relationships. It uses a masked autoencoder architecture to enforce sparsity and incorporates a differentiable acyclicity condition to ensure that the resulting graph is a valid DAG \cite{bello2022dagma}.
    \item \textbf{LLM-based Pairwise Method}: This approach utilizes large language models to infer causal relationships through pairwise queries based on metadata and variable descriptions. Although intuitive and effective for small graphs, it scales poorly, requiring $\mathcal{O}(n^2)$ queries for $n$ variables \cite{kiciman2023causal}.
    \item \textbf{LLM-based BFS Method}: The performance of the proposed method is also compared to that of the LLM-based BFS appraoch introduced by Jiralerspong et al. \cite{jiralerspong2024efficient} which was discussed in the \ref{sec:introduction} and \ref{sec:Method} sections. The implementations utilized for all the baseline methods are from the repository provided by the authors of \cite{jiralerspong2024efficient}.
    
\end{itemize}

\subsection{Fairness Analysis through Pathway Analysis}

The fairness analysis conducted in this work evaluates both direct and indirect effects, providing insights into fairness-related dynamics. To assess the contribution of the fairness-related paths and distinguish the relative contributions of direct and indirect pathways, we reported the following:

\begin{enumerate}
    \item The total number of direct paths and indirect paths from the sensitives attribute to the target variable.

    \item The effects of sensitive variables, decomposed into:
    \begin{itemize}
        \item \textbf{Direct Effect (DE):} The causal effect of $S$ on $Y$ through direct pathways.
        \item \textbf{Indirect Effect (IE):} The causal effect of $S$ on $Y$ through intermediate variables.
        \item \textbf{Total Effect (TE):} The sum of $DE$ and $IE$:
        $
        TE = DE + IE
        $
    \end{itemize}

    \item To understand bias dynamics, these effects are normalized with respect to the total outcome variance.

    \begin{equation}
        C_{\text{bias}} = \frac{TE}{\text{Total Outcome Variance}}
    \end{equation}
    
\end{enumerate}

Using the Adult UCI dataset, where fairness concerns are prominent, sensitive variables (Sex, Age, Race), outcome variables (Salary), and intermediate variables are identified. Each causal discovery method generates (1) Causal Graphs: Represented by adjacency matrices capturing the graph structure, and (2) Causal Dictionaries: Capturing identified causal relationships. These outputs are analyzed to compute the above mentioned fairness-related metrics,

\section{Results and Analyses}

\subsection{Parameter Impact Analysis on Proposed Method}

To evaluate the influence of various parameters on the performance of the proposed method, a detailed analysis of the hyperparameters is conducted using visualizations such as ridge plots, correlation heatmaps, and feature importance analysis.


\subsubsection{Impact of Weights on Performance}

\begin{figure}[h]
    \centering
    \begin{minipage}{0.45\textwidth}
        \includegraphics[width=\textwidth]{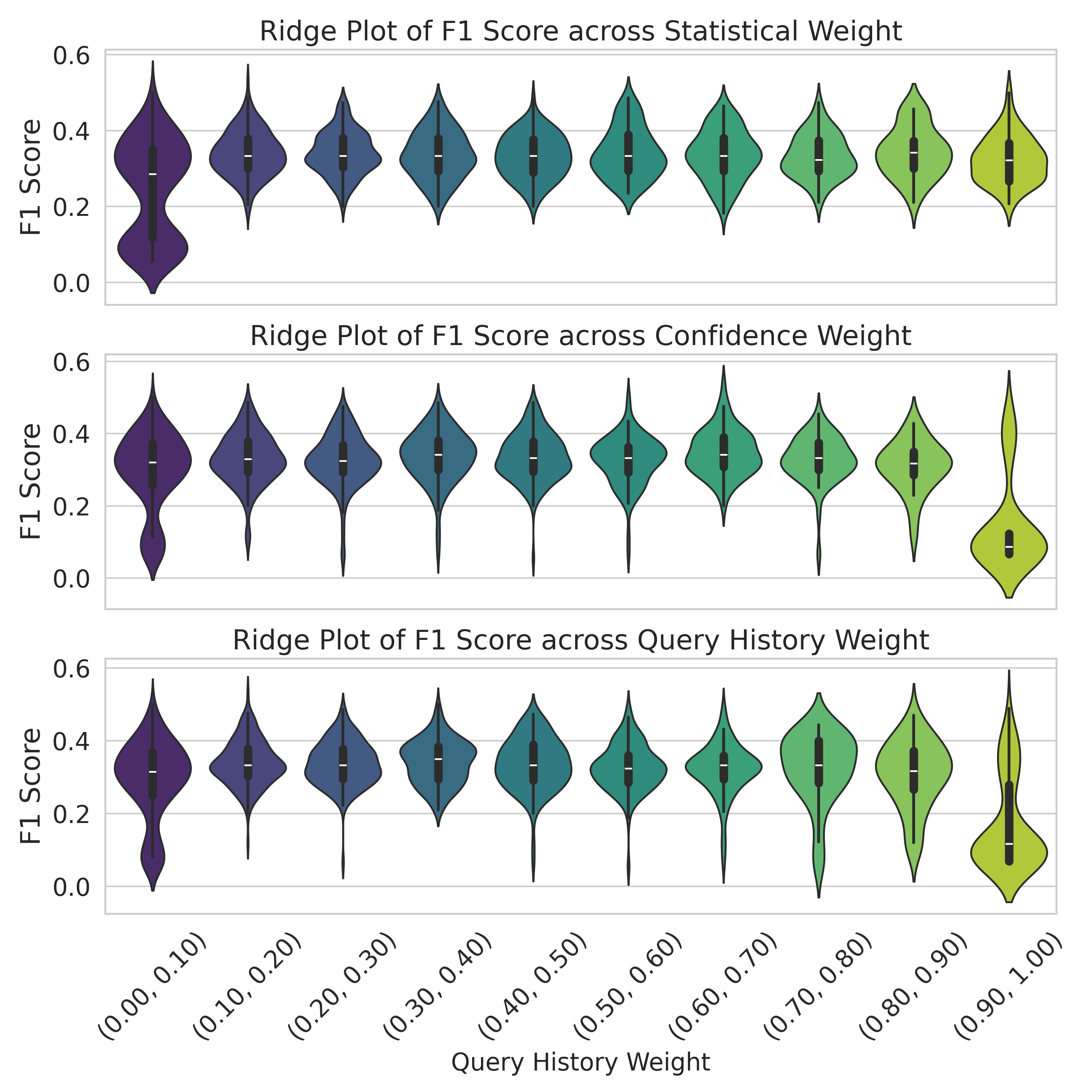}
        \caption{Ridge Plots for Weights of Statistical Information, LLM Confidence, and Query History against F1-Score}
        \label{fig:ridge_plots}
    \end{minipage}
    \hfill
    \begin{minipage}{0.45\textwidth}
        \includegraphics[width=\textwidth]{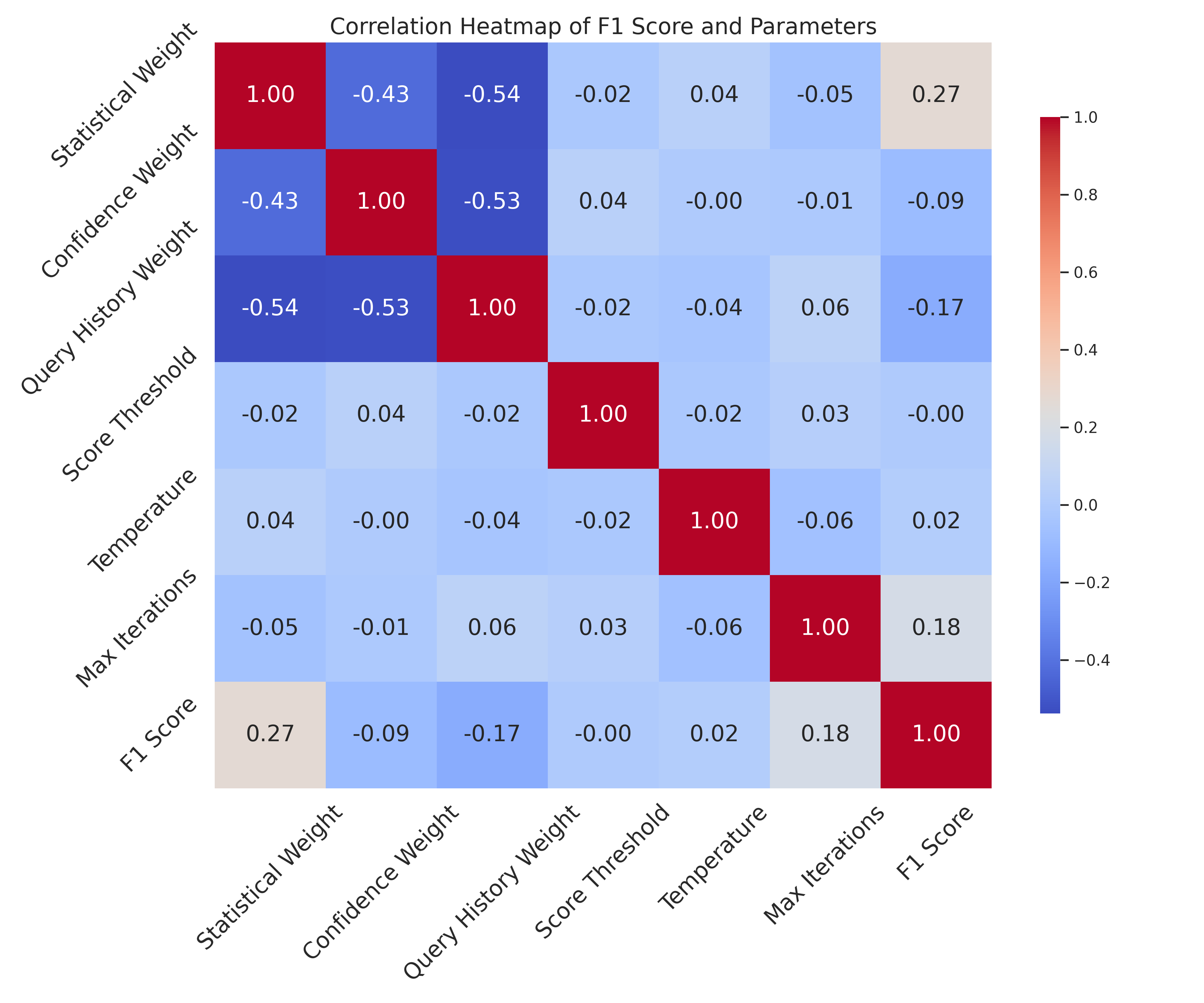}
        \caption{Correlation Heatmap of Parameters and F1 Score}
        \label{fig:corr_map}
    \end{minipage}
\end{figure}

Figure \ref{fig:ridge_plots} shows ridge plots illustrating the impact of the 3-weight combination on the F1 score for experiments on the Child network. Each plot visualizes F1 score distributions across different weight bins. The width indicates variation in F1 scores; a wider ridge means more variation, while a narrower ridge suggests consistent performance. The height of the peak shows the most frequently occurring F1 scores, with a taller peak indicating a common F1 score for that weight bin. Colors distinguish ridges for each bin without indicating performance or significance. The figure shows that statistical information and query history are significant contributors to performance, while LLM confidence has a stable, less pronounced effect.

\begin{itemize}
    \item Statistical Information Weight: Lower bins (0.0-0.1) show narrower ridges and lower F1 scores. As the weight increases (0.3-0.6), distributions broaden, and peaks shift toward higher F1 scores, suggesting optimal performance in this range. Beyond 0.6, performance flattens with diminishing returns.

\item LLM Confidence Weight: Distributions remain stable across all bins, with minor improvements in midrange bins (0.3-0.5). This weight has a minor impact on the F1 score and serves as a supporting factor rather than a decisive one.

\item Query History Weight: Lower bins (0.0-0.2) show lower F1 scores. As the weight increases (0.4-0.6), distributions widen, and peaks shift toward higher F1 scores. Beyond 0.6, improvement plateaus, with optimal performance around 0.4-0.6.
\end{itemize}

These results emphasize the importance of balancing weights, as increasing one decreases the others due to the constraint that all weights must sum to 1.



\subsubsection{Correlation Analysis}

Figure \ref{fig:corr_map} illustrates the relationships between model parameters and the F1 score, revealing key performance factors. The weight of statistical information has the strongest positive correlation with the F1 score (+0.27), indicating that increasing its value generally improves performance. The weight of query history shows a weaker negative correlation (-0.17), reflecting diminishing returns beyond its optimal range (0.4-0.6). The weight of the LLM confidence score has a minimal impact (-0.09 correlation), therefore its optimization is less critical. Negative correlations between weights (e.g., -0.54 between statistical information and query history) highlight trade-offs due to their sum equaling 1. The score threshold shows negligible influence on the F1 score and can be deprioritized. Temperature and the number of maximum iterations have weak positive correlations (+0.02 and +0.18, respectively), so small adjustments have limited impact.

These findings underscore the importance of optimizing statistical information and query history weights, while treating other parameters as secondary contributors to model performance.


\subsection{Comparison with Baseline Methods}


As seen in Table \ref{tb:performance}, the proposed method outperformed the baseline methods in most evaluation metrics. Importantly, we compared the best-performing results of the proposed method with those of each baseline method, as performance differences between different runs within each method were not significant.

For the Child network, the proposed method achieved the highest accuracy (0.364) among all methods and precision (0.6), outperforming GES (0.438). It also showed the second highest recall (0.48), slightly below the LLM pairwise method (0.56) and resulted in the highest F1 score (0.533). This indicates a strong balance in predicting true edges while minimizing false positives and negatives. Both the true and estimated graphs adhered to the DAG constraints. The proposed method achieved a NHD of 0.082, slightly higher than NOTEARS (0.080) but better than all other methods. The lowest ratio of normalized NHD to reference NHD (0.467) highlights its superior performance.

The runtime of the proposed method for Child (25.37 seconds) was higher compared to PC (0.12 seconds), GES (9.47 seconds) and NOTEARS (5.39 seconds). This is expected for smaller networks like the "Child" graph, where statistical methods are faster. However, the proposed method, with its optimization techniques and LLM-based querying, scales effectively for larger and complex networks, as seen in the results for the Neuropathic network. This trade-off highlights the balance between runtime and robustness.

For the Neuropathic network, the proposed method achieved the highest F1 score (0.136) among all methods. The precision (0.690) of the proposed method was significantly higher than that of all other methods, while the recall (0.075) was comparable to other methods.  This disparity reflects the trade-off between achieving high precision and covering all true edges in large networks. The NHD of 0.109 was the second lowest, while it achieved the lowest ratio of normalized NHD to reference NHD, suggesting stronger graph reconstruction performance compared to the other methods. Accuracy (0.073) and the number of predicted edges (84) were higher than most baselines.

The proposed method completed the Neuropathic experiment in 94.24 seconds faster than DAGMA (4196.85 secs) and NOTEARS (1896.16 secs). This efficiency demonstrates the scalability of the proposed approach, particularly when applied to complex networks. While the runtime is slightly higher than PC (88.90 secs), the proposed method provided substantially better performance across all metrics, justifying this trade-off.

The LLM BFS method, however, achieved 0 accuracy and F1 score for the Neuropathic network, which may indicate limitations in its BFS query strategy when applied to highly complex and large-scale networks. Specifically, the high dimensionality and sparsity of the Neuropathic graph might have overwhelmed the query prioritization mechanisms. This highlights a key challenge for methods that rely heavily on LLM-based evaluations or search strategies in large graphs, where their performance depends on efficient query selection and graph traversal strategies. The performance of the proposed method illustrates how enhancements, such as dynamic scoring and active learning, address these issues by refining query prioritization to better handle complexity and scale.

Due to limitations in resources, experiments using the GES and LLM pairwise methods were not conducted on the Neuropathic graph. 

\vskip -0.1in

\begin{table*}[h]
\caption{Performances of Methods on the Child and Neuropathic Causal Networks}
\label{tb:performance}
\begin{center}
\begin{scriptsize}
\begin{tabular}{l||l||ccccccc|c}
\toprule
   \textbf{Child} &\textbf{Method} & \textbf{Acc. ($\uparrow$)} & \textbf{F Score ($\uparrow$)} & \textbf{Precision} & \textbf{Recall} & \textbf{NHD}  & \textbf{Ratio ($\downarrow$)} & \textbf{predicted Edges} & \textbf{Runtime (s) ($\downarrow$)} \\
   \cmidrule{2-10}

  (20 nodes, 25 edges) & PC & 0.146 & 0.255 & 0.273 & 0.239 & 0.097 & 0.745 & 22 & 0.12\\
  & GES & 0.206 & 0.341 &  0.438 & 0.279 & 0.119 & 0.659 & 16  & 9.47 \\
  & NOTEARS & 0.216 & 0.356 & 0.403 & 0.319 & 0.080 & 0.644 & 20 & 5.39\\
  & DAGMA $(\lambda = 0.01)$ & 0.179 & 0.304 & 0.333 & 0.279 & 0.089 & 0.696 & 21 & 1.71\\
  & LLM Pairwise & 0.130 & 0.229 & 0.144 & \textbf{0.559} & \textbf{0.235} & 0.770 & 97 & 1461.78 \\
  & LLM BFS & 0.150 & 0.261 & 0.286 & 0.240 & 0.085 & 0.739 & 21 & 22.77\\

\cmidrule{2-10}

    & Proposed Method & \textbf{0.364} & \textbf{0.533} & \textbf{0.601} & 0.479 & 0.082 & \textbf{0.467} & 20 & 25.37 \\

\midrule

  \textbf{Neuropathic} & PC & 0.041 & 0.078 & 0.092 & 0.068 & \textbf{0.025} & 0.922 & 563 & 88.90\\
  (221 nodes, 770 edges) & GES & N/A & N/A & N/A & N/A & N/A & N/A & N/A & N/A \\
  & NOTEARS $(\lambda = 0.01)$ & 0.022 & 0.044 & 0.5 & 0.023 & 0.334 & 0.955 & 36 & 1896.16 \\
  & DAGMA $(\lambda = 0.01)$ & 0.020 & 0.039 & 0.421 & 0.021 & 0.351 & 0.960 & 38 & 4196.85\\
  & LLM Pairwise & N/A & N/A & N/A & N/A & N/A & N/A & N/A & N/A \\
  & LLM BFS & 0.00 & 0.00 & 0.00 & 0.00 & 0.903 & 1.000 & 43 & 42.22 \\

\cmidrule{2-10}
    & Proposed Method & \textbf{0.073} & \textbf{0.136} & \textbf{0.690} & \textbf{0.075} & 0.109 & \textbf{0.864} & 84 & 94.24 \\

\bottomrule
\end{tabular}
\end{scriptsize}
\end{center}
\vskip -0.1in
\end{table*}

\subsection{Fairness Analyses Across Methods on ADULT Dataset}

The fairness analysis of different CD methods reveals significant variability in how sensitive attributes (Age, Sex, Race) influence the outcome variable (Salary), both directly and indirectly. Table \ref{tb:fairness_results} summarizes key metrics like the number of direct and indirect paths, their effects, the total effect, and the normalized bias contribution ($C_{\text{bias}}$).

\begin{itemize}
    
\item PC Algorithm: Identified 1 direct path and 32 indirect paths, with a Total Effect (TE) of 0.3861 and a high $C_{\text{bias}}$ of 2.0713. It captures mediated influences but is prone to spurious paths.

\item GES: Found 2 direct paths and 5 indirect paths, with a TE of 0.2142 and a $C_{\text{bias}}$ of 1.1490. The focus is on direct paths, offering a concise causal structure.

\item Pairwise LLM: Discovered 3 direct paths and 401 indirect paths, with the highest TE of 0.3892 and a $C_{\text{bias}}$ of 2.0878. Extensive mediated relationships.

\item BFS-based LLM: Found no direct paths, a TE of 0.0001, and minimal $C_{\text{bias}}$. Its utility for fairness analyses is limited.

\item Proposed Method: Provided a streamlined causal graph with 1 direct path, a low TE of 0.0077, and a $C_{\text{bias}}$ of 0.0415. This emphasizes simplicity but misses key variables like race.

\end{itemize}

This analysis shows that different CD methods produce varying causal graphs, affecting fairness interpretation. Methods like PC and LLM Pairwise capture extensive indirect pathways but introduce noise. GES and the proposed method focus on direct paths, resulting in simpler but possibly incomplete representations. These results stress the importance of selecting appropriate methods tailored to fairness evaluation objectives.

\begin{table*}[h]
\caption{Fairness Analysis Results Across Different CD Methods}
\label{tb:fairness_results}
\begin{center}
\begin{scriptsize}
\begin{tabular}{|l||c|c|c|c|c|c|}
\hline
\textbf{Method} & \textbf{Direct Paths} & \textbf{Indirect Paths} & \textbf{Direct Effect (DE)} & \textbf{Indirect Effect (IE)} & \textbf{Total Effect (TE)} & \textbf{Normalized Bias Contribution} \\ 
\hline
PC & 1 & 32 & 0.1989 & 0.1872 & 0.3861 & 2.0713 \\ 
\hline
GES & 2 & 5 & 0.2066 & 0.0075 & 0.2142 & 1.1490 \\ 
\hline
LLM Pairwise & 3 & 401 & 0.2436 & 0.1457 & 0.3892 & 2.0878 \\ 
\hline
LLM BFS & 0 & 11 & 0.00 & 0.0001 & 0.0001 & 0.0006 \\ 
\hline
Proposed Method & 1 & 0 & 0.0077 & 0.00 & 0.0077 & 0.0415 \\ 
\hline
\end{tabular}
\end{scriptsize}
\end{center}
\vskip -0.1in
\end{table*}


\section{Discussion}

The results highlight the effectiveness and versatility of the proposed method, particularly in improving CD and fairness analysis using LLM. This discussion will address key findings from parameter impact analysis, baseline comparisons, and fairness evaluations while contextualizing the broader implications for the field.

\subsection{Parameter Impact Analysis}

The parameter impact analysis demonstrates that careful tuning of weights for statistical information and query history can substantially enhance the model's performance. The ridge plot analysis shows statistical information greatly improves the F1 score, while query history, though influential, has diminishing returns beyond certain thresholds. Conversely, the LLM confidence weight exhibits limited impact, suggesting that this parameter plays a supporting role rather than being a decisive factor. This result shows the importance of prioritizing the tuning of parameters that directly influence query informativeness, which aligns with the AL paradigm's goal of maximizing efficiency.

The findings further emphasize the trade-offs in balancing these weights, as shown by negative correlations between them,  must be managed carefully to optimize the overall performance of the CD process while maintaining consistency and robustness across different datasets. Similar correlation patterns were observed in the Neuropathic experiments; however, the correlation between the F1 score and the maximum number of iterations was notably higher in those experiments. This observation aligns with the relatively low recall achieved by the proposed method on Neuropathic, suggesting the possibility that not all true causal edges were identified. This limitation highlights the need to optimize the recall-focused components of the dynamic scoring mechanism, such as query prioritization and termination thresholds.

\subsection{Baseline Comparisons}

The proposed method outperformed baseline methods in nearly all metrics for the Child dataset. It achieved the highest F1 score and precision, with a competitive recall, lower normalized Hamming distance (NHD), and the best NHD-to-reference ratio, indicating better structural alignment with the ground-truth graph.

Importantly, the improvements come at the cost of higher runtime compared to statistical methods like PC and GES. However, this trade-off is justified for larger and more complex causal networks such as Neuropathic where statistical methods were shown to face scalability issues. The LLM Pairwise method, although capable of capturing extensive indirect pathways, suffers from quadratic scaling of queries and lower precision due to its exhaustive approach. The proposed method, which uses AL and dynamic scoring, addresses these limitations, making it a more practical and efficient choice for real-world applications.

\subsection{Fairness Analysis}

The fairness analysis revealed variability in how different CD methods interpret the influence of sensitive variables on outcomes, supporting the findings of Binkyte et al. \cite{binkyte2023causal}. This study gives further insights into how LLM-based causal methods differ in this aspect. Statistical methods like PC capture a larger number of indirect paths, reflecting their ability to identify mediated effects. However, this comes at the risk of introducing spurious pathways, as evidenced by their higher normalized bias contribution. In contrast, LLM-based methods prioritize direct and interpretable paths, with the proposed method striking a balance by identifying direct relationships while avoiding excessive complexity.

Interestingly, the proposed method produced a streamlined causal graph with minimal indirect paths and the lowest normalized bias contribution. While this simplicity enhances interpretability and reduces spurious paths, it also raises concerns about missing critical fairness-relevant variables, as evidenced by its relatively low total effect (TE). This highlights the need for careful consideration of fairness objectives when selecting CD methods. For scenarios where comprehensive mediated pathways are essential, methods like PC or LLM Pairwise may be preferable. However, for applications requiring concise and interpretable graphs, the proposed method offers a compelling alternative.

\subsection{Limitations}

Although the proposed method advances LLM-based causal discovery, it has certain limitations. It heavily depends on the underlying LLM, leading to performance variability due to differences in model architecture, pre-training data, and internal reasoning processes. Variations across LLM versions or deployments may cause inconsistent results. The method involves multiple hyperparameters (e.g., weights for dynamic scoring, query thresholds, and LLM temperature), requiring significant tuning effort and domain knowledge. Additionally, LLM-based methods incur substantial computational costs, especially for large datasets, posing a scalability barrier for researchers with limited resources.

Current LLMs impose strict token limits per query. For networks with extensive variable definitions, like the Neuropathic graph, this constraint prevents the pass of the full metadata, leading to incomplete responses. This limitation affects the quality of inferred causal graphs, as seen in Neuropathic network experiments. The LLM occasionally introduces typos or variations in variable names, causing errors in the predicted graph. High temperature settings, encouraging creative outputs, can lead to such inaccuracies and additional debugging challenges.

These limitations highlight areas for future improvement, including standardizing LLM outputs, handling large metadata more effectively, and mitigating the computational overhead of LLM-based querying.

\section{Conclusion}

This study further offers valuable insights into optimizing LLMs for CD and their potential in fairness evaluation. By analyzing parameter effects, it advances understanding of efficient LLM use for constructing accurate and interpretable causal graphs. The results show the impact of improved causal graph accuracy on fairness analysis, emphasizing the need for further research on CD methods to address bias and ensure fairness in machine learning applications.

The proposed method's AL framework and dynamic scoring mechanism pave the way for scalable and efficient LLM-based CD approaches. The proposed method significantly reduces query complexity while maintaining performance in terms of precision, F1 score, and graph reconstruction quality. Its scalability to larger networks demonstrates its potential for real-world applications, where high-dimensional and sparse graphs are common. Experiments showed that while the method incurs higher runtimes for smaller graphs, its ability to handle larger, more complex networks effectively compensates for this trade-off. As LLMs improve, future work can explore integrating additional domain knowledge, enhancing scalability, and refining methods to balance runtime with accuracy, especially for larger and more complex datasets.





\newpage



\newpage
\appendix

\section{Details about Baseline Methods} \label{apdx:baselines}

\subsection{Peter-Clark (PC) Algorithm}

The PC algorithm is a constraint-based approach to discover causal structures \cite{spirtes1991algorithm}. It starts with a fully connected graph and applies conditional independence tests to remove edges, creating a graph skeleton. The algorithm then orients edges to produce a DAG by identifying v-structures and ensuring acyclicity. The PC algorithm assumes causal sufficiency and faithfulness, relying on accurate conditional independence testing.

\subsection{Greedy Equivalence Search (GES) Algorithm}

The GES algorithm is a score-based method that learns causal structures by optimizing a predefined score function like BIC \cite{meek1997graphical}. It has two phases: forward, adding edges to improve the score, and backward, removing edges for further improvement. GES works with equivalence classes of DAGs, avoiding redundant structures. It assumes causal sufficiency and may not always find the global optimum.

\subsection{NOTEARS Algorithm}

The NOTEARS algorithm formulates causal structure learning as a continuous optimization problem \cite{zheng2018dags}. It uses a smooth acyclicity constraint based on the adjacency matrix, allowing gradient-based optimization. NOTEARS balances data fit and model sparsity, assuming causal sufficiency and requiring appropriate loss functions and regularization.

\subsection{DAG Learning with Masked Autoencoders (DAGMA) Algorithm}

DAGMA integrates neural networks and optimization to learn causal structures \cite{bello2022dagma}. It ensures acyclicity through a differentiable function and models complex, non-linear dependencies. DAGMA uses a masked autoencoder to encourage sparsity and optimize the graph structure and model parameters simultaneously, suitable for high-dimensional data.

\subsection{LLM-based Pairwise Method}

Previously introduced LLM-based CD methods utilize pairwise queries to infer the causal relationship between two variables at a time, using metadata associated with the input variables. This approach mimics the process by which human experts construct causal graphs using domain knowledge, similar to the proposed method.

Each query determines whether changing one variable affects another. For two variables, $A$ and $B$, the possible causal relationships are: $A \rightarrow B$, $A \leftarrow B$, or no relationship. The LLM processes a natural language prompt like:

\begin{quote}
\texttt{<A>: <Description A>}\\
\texttt{<B>: <Description B>}\\
\texttt{Given the above information, which of the following is the most likely:}\\
\texttt{A. Changing <A> causes a change in <B>}\\
\texttt{B. Changing <B> causes a change in <A>}\\
\texttt{C. There is no causal relationship between <A> and <B>}
\end{quote}

The LLM selects the most likely relationship based on its internal reasoning. Constructing a full causal graph requires $\mathcal{O}(n^2)$ queries for $n$ variables. This method, demonstrated by Kıcıman et al. \cite{kiciman2023causal}, scales poorly and is limited to small causal graphs due to high computational expense.
This LLM-based pairwise approach serves as a baseline for evaluating methods that aim to incorporate domain knowledge in CD tasks, particularly in cases where observational data are unavailable.

\end{document}